\documentclass{article}

\long\def\ignore#1{}

\usepackage[final]{neurips_2021}

\usepackage[utf8]{inputenc} % allow utf-8 input
\usepackage[T1]{fontenc}    % use 8-bit T1 fonts
\usepackage{hyperref}       % hyperlinks
\usepackage{url}            % simple URL typesetting
\usepackage{booktabs}       % professional-quality tables
\usepackage{amsfonts}       % blackboard math symbols
\usepackage{nicefrac}       % compact symbols for 1/2, etc.
\usepackage{microtype}      % microtypography
\usepackage{xcolor}         % colors
\usepackage{ulem}           % strikethrough
\usepackage[colorinlistoftodos]{todonotes} % todo comments
\usepackage{natbib}         % Natbib citation styles

\setcitestyle{authoryear,open={(},close={)}} %Citation-related commands
\title{EAANet: Efficient Attention Augmented Convolutional Networks}
\author{%
  Runqing Zhang \\
  Department of Computer Science\\
  University of Toronto\\
  \texttt{runqingz@cs.toronto.edu} \\
  \And
  Tianshu Zhu \\
  Department of Computer Science\\
  University of Toronto\\
  \texttt{tianshu@cs.toronto.edu} \\
}

\begin{document}

\maketitle

\maketitle

\begin{abstract}
Humans can effectively find salient regions in complex scenes. Self-attention mechanisms were introduced into Computer Vision (CV) to achieve this. Attention Augmented Convolutional Network (AANet) is a mixture of convolution and self-attention, which increases the accuracy of a typical ResNet. However, The complexity of self-attention is $\mathcal{O}(n^2)$ in terms of computation and memory usage with respect to the number of input tokens. In this project. We propose EAANet: Efficient Attention Augmented Convolutional Networks, which incorporates efficient self-attention mechanisms in a convolution and self-attention hybrid architecture to reduce the model’s memory footprint. Our best model show performance improvement over AA-Net and ResNet18. We also explore different methods to augment Convolutional Network with self-attention mechanisms and show the difficulty of training those methods compared to ResNet. Finally, we show that augmenting efficient self-attention mechanisms with ResNet scales better with input size than normal self-attention mechanisms. Therefore, our EAANet is more capable of working with high-resolution images.
\end{abstract}

\section{Introduction}
%talk about CNN and self-attention
Convolutional Neural Networks (CNNs) have succeeded in computer vision, especially in image classification. The convolutional layer's design forces locality through a small receptive field. The convolutional layer also forces translation equivariance through weight sharing. When designing models that operate over images, both properties prove to be critical inductive biases. However, the convolutional kernel's local nature prevents it from capturing global contexts in an image, which is often required for better object recognition in images \citep{rabinovich2007objects}. Self-attention \citep{vaswani2017attention}, on the other hand, has emerged as a breakthrough in capturing long-range interactions, but it has primarily been used in sequence modeling and generative modeling tasks. The main idea behind self-attention is to compute a weighted average of hidden unit values. Unlike the pooling or convolutional operators, the weights used in the weighted average operation are generated dynamically using a similarity function between hidden units. As a result,  the interaction between input signals is determined by the signals rather than their relative location. This result enables self-attention to capture long-range interactions.

%talk about AANet and EAANet
As an alternative to convolutions, the AANet proposed by \citet{bello2019attention} considered the use of self-attention for discriminative visual tasks. They proposed using this self-attention mechanism to improve convolutions. Convolutional feature maps, which enforce locality, are combined with self-attentional feature maps, which can model longer-range dependencies. However, the computation and memory complexity of self-attention layers scales quadratically with input size. This problem is more severe with AANet because the image data is in 2-D, as opposed to natural language tasks where the data is in 1-D. We incorporate a 2-D version of efficient self-attention mechanism called Efficient Vision Transformer (E-ViT) proposed by \citet{zhang2021multi} into a ResNet18 \citep{he2016deep} in this project to alleviate high memory stress when a AANet works with high-resolution images.
%talk about our main contributions

Our main contributions are as follows:
\begin{itemize}
    \item We present EAANet, a hybrid architecture of augmented ResNet and efficient self-attention mechanisms.
    \item We evaluate the performance of our model on CIFAR-10. Our model has a noticeable performance gain over AANet and ResNet.
    \item We extend EAANet by evaluating different augmentation methods and efficient self-attention mechanisms, specifically, Linformer and Longformer.
    \item We explore the training complexity of EAANet, and we show that our model scales better with input size than AANet.
\end{itemize}

\section{Related Work}
\subsection{Convolution and Self-attention Mechanisms}
Recent work in Natural Language Processing has explored the use of self-attention in conjunction with convolutions. Because of its ability to capture long-distance interactions, attention has become a popular computational module for modeling sequences. \citet{bahdanau2014neural}, for example, was the first to propose combining attention with a Recurrent Neural Network \citep{hochreiter1997long} for Machine Translation alignment. \citet{vaswani2017attention} extended attention even further, achieving state-of-the-art results in Machine Translation with the self-attentional Transformer architecture.

The current success of the self-attention mechanism in NLP tasks has resulted in many CV adoptions. Given that self-attention will capture the image's global information, many scientists have combined self-attention and convolution to boost CV task performance. Squeeze-and-Excitation \citep{hu2018squeeze} and Gather-Excite \citep{hu2018gather}, for example, use signals aggregated from entire feature maps to reweigh feature channels. The Block Attention Module (BAM) \citep{park2018bam} and the Convolutional Block Attention Module (CBAM) \citep{woo2018cbam} infer attention maps along the channel and spatial dimensions of a given feature map in a sequential manner. These attention maps will be multiplied back to the feature map for refinement. The self-attention map is directly concatenated with the convolution result in the Attention Augmented Convolutional Network (AAnet) \citep{bello2019attention} to ensure that both locality information and long-range dependencies are captured. Multi-head attention allows the model to pay attention to both spatial and feature subspaces simultaneously. \citet{bello2019attention} also improved the representational power of self-attention over images by extending 1-D relative self-attention \citep{shaw2018self} to 2-D inputs, allowing them to model translation equivariance in a principled way.

\subsection{Efficient Self-attention Mechanisms}
Modern Computer Vision has been built on powerful image embedding learned on image classification tasks such as CIFAR-10 \citep{cifar10}, CIFAR-100 \citep{cifar100}, and ImageNet \citep{deng2009imagenet}. These datasets have been used as benchmarks for training better image embedding and network architectures across many tasks. Developing an efficient self-attention mechanism for high-resolution image embedding is crucial in CV. A well-known deficiency of the self-attention mechanism is its quadratic computation time and memory complexity. Many efficient attention mechanisms have been developed for Transformers to reduce the computation and memory complexity, most for NLP tasks. These mechanisms can be grouped into four categories \citep{zhang2021multi}. 

\begin{itemize}
    \item [1.] Sparse attention mechanism, including content-independent sparsity and content-dependent sparsity. Axial Transformer \citep{ho2019axial} and Image Transformer \citep{parmar2018image} are among few sparsity-based efficient attentions that are developed for image generation.
    \item [2.] Memory-based mechanism, including Compressive Transformers \citep{rae2019compressive} and Set Transformer \citep{lee2019set}. These models use some extra global tokens as static memory and allow all the other tokens to attend only to those global tokens.
    \item [3.] Low-rank-based mechanism. For example the Linformer \citep{wang2020linformer} reduces the overall self-attention complexity from $\mathcal{O}(n^2)$ to $\mathcal{O}(n)$ by projecting key $K$ and value $V$ with $N$ tokens to a low-dimensional representation. This low-rank approximation of the length dimension will decompose the original self-attention matrix from $N\times N$ to $N \times K$. Similar to Linformer, the Spatial Reduction Attention (SRA) presented in \citep{wang2021pyramid} also conduct projection on key and value but using convolution layer with kernel size $R$ and stride $R$.
    \item [4.] Kernel-based mechanism like Performer \citep{choromanski2020rethinking}.
\end{itemize}

Longformer \citep{beltagy2020longformer} utilizes hybrid attention mechanisms. It combines the sparsity and memory-based mechanism with its unique sliding window approach during attention score calculation. Compared with vanilla self-attention, which attends all tokens in every iteration, Longformer has a fixed interval of tokens to reduce the computation expenses. This sliding window attention mechanism can be improved by adding some global tokens that attend every iteration.

Multi-scale Vision Transformer \citep{zhang2021multi}, to our knowledge, is the first work implementing efficient Transformers such as Longformer and Linformer in computer vision tasks. A comprehensive evaluation of different efficient Transformer models' performance is also presented in Multi-scale Vision Transformer repository. We also use the efficient Transformers implementation in Multi-scale Vision Transformer of Longformer and Linformer blocks in our EAANet.

\section{Efficient Attention Augmentation Network}

Like AANet, our work augments ResNet with E-ViT blocks implemented in Multi-Scale Vision Transformer. We explore replacement or concatenation when augmenting E-ViT with ResNet. The replacement approach can either replace a single convolution operation or a Residual block (a group of convolution operations wrapped in residual connection). The concatenation approach only adds E-ViT block to each Residual block. During the forward or backward path of the network, the concatenation approach serves as auxiliary information flow to Residual connection. In addition, we think concatenating every convolution operation over complicates the model.

We use the same cross-entropy loss as ResNet since our augmentation does not change the output dimension. ResNet defines four layers Residual blocks with the same channel size within each layer. Multi-scale Vision Transformer also defines four different E-ViT blocks. Intuitively, we augment Residual blocks and E-ViT blocks at the same level. Figure 1 gives a visual representation of our overall model.
 
\begin{figure}[ht]
  \centering
  \includegraphics[scale=0.3]{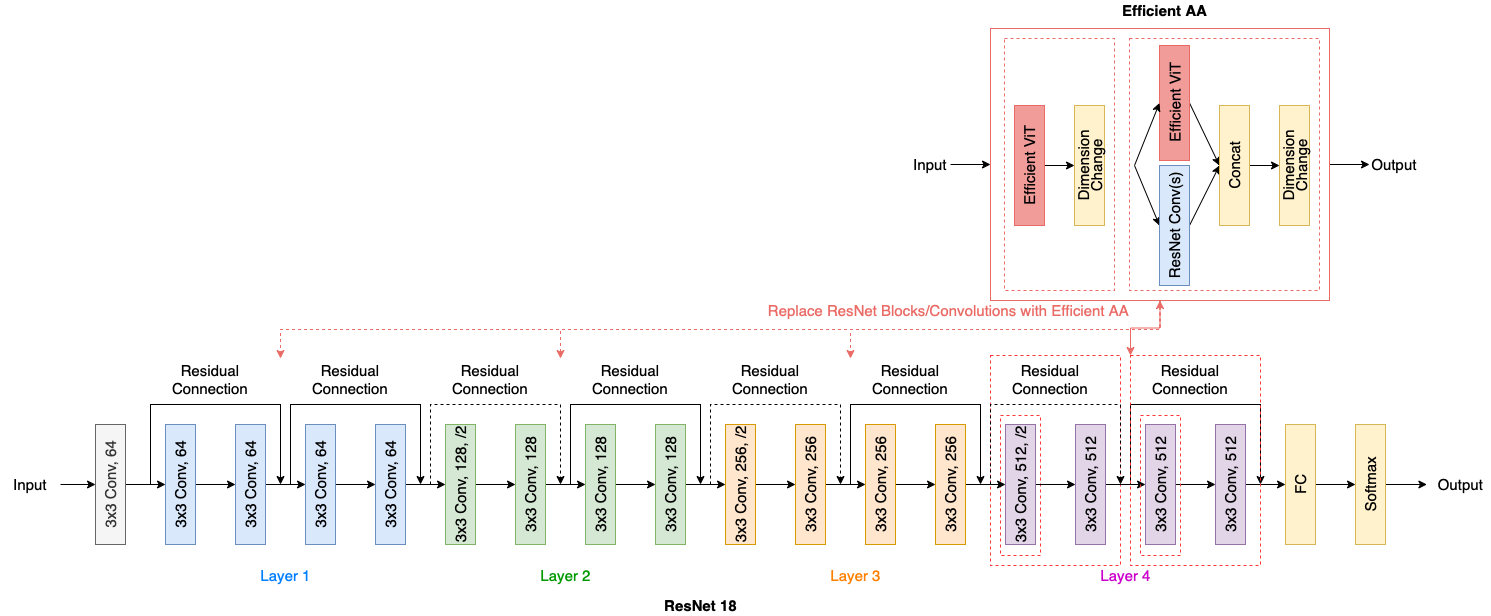}
  \caption{Similar to AANet that can be used in many backbone models. We based our EAANet on ResNet18. We replaced residual blocks/convolutional layers with Efficient Attention Augmented Convolution (EAA).}
\end{figure}

\subsection{Efficient Attention Augmentation}

\begin{figure}[ht]
  \centering
  \includegraphics[scale=0.5]{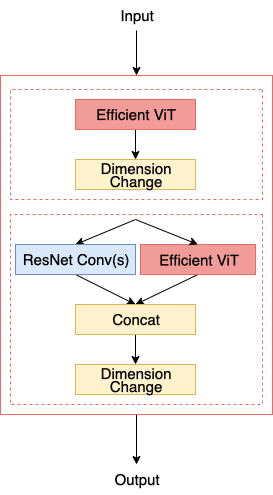}
  \caption{The two approaches of EAANet. First is to concatenate the computed output from both Residual block and Efficient-ViT block. The second is to replace convolution(s) completely with Efficient ViT Block.}
\end{figure}

Our two approaches to Efficient Attention Augmentation are shown in Figure 2. The first approach is to concatenate the output from the Residual block (which contains multiple convolution operations $G$) and E-ViT block, $E$. Then we do a linear projection $F$ to make sure the output channel size matches the next Residual block input channel size:

\begin{equation}
    Y = F(Concat((G(X) + X), E(X)))
\end{equation}

The second approach is to replace a single convolution operation or a Residual block with E-ViT block, $E$, then perform a linear projection $F$ to make sure the output dimension matches the next convolution operation input:

\begin{equation}
    Y = F(E(X))
\end{equation}

The first Residual block of each ResNet layer starting from layer 2 downsamples the image size by 2 with a stride of 2 convolution. Intuitively, we can add another stride of 2 convolution and use 1x1 image patches in E-ViT similar to AANet. We also explore directly using 2x2 image patches E-ViT to reduce the image size by 2. Empirically, using 2x2 image patches performs better, as shown in the experiment section of this report.

\section{Experiments}
\subsection{CIFAR-10}
We evaluated our models’ performance on CIFAR-10 dataset \citep{cifar10}. We present the performance result against ResNet18 in Table 1. Overall, we find out that attention concatenation significantly improves the performance of ResNet on CIFAR-10. We think this is mainly because by concatenating attention to ResNet, we have the information from both attention and convolution, which is better than vanilla ResNet with only convolution information.

\begin{table}[t]
  \centering
  \begin{tabular}{lll}
    \toprule
    Model     & Top1 Accuracy ($\%$)     & Top5 Accuracy ($\%$) \\
    \midrule
    ResNet18 & 90.83  & 99.48     \\
    AA-Net     & 89.5 & 99.37      \\
    EAANET-Replacement     & 89.86       & 99.46  \\
    EAANET-Concatenation     & 92.52       & 99.68  \\
    \bottomrule
  \end{tabular}
  \caption{We compare the performance between ResNet18, AA-Net Based on ResNet18, replacing ResNet Convolution block with Efficient Transformers block (EAANET-Replacement), and concatenating Efficient Transformer block to ResNet block (EAANET-Concatenation). AA-Block and EAANET-Replacement models show similar performance compared to ResNet. EAANET-Concatenation models show a $3\%$ improvement over ResNet.}
\end{table}

\subsection{Architecture}
\subsubsection{Efficient Transformer Type}
There are many efficient self-attention mechanism implementations. In our work, we specifically pick Longformer and Linformer, which show the best performance in the Multi-scale Vision Transformer model\citep{zhang2021multi}. In Table 2, we show our test results. We find that for either replacement or concatenation approach, using Longformer performs better than using Linformer. We think this is mainly because Linformer uses a lower rank approximation which harms the final accuracy \citep{wang2020linformer}. Longformer, on the other hand, calculates attention scores in a local neighborhood without approximation \citep{beltagy2020longformer}.

One major issue with using Longformer is the extremely slow processing time. For similar model size (small version defined in Multi-scale Vision Transformer model) and input size (CIFAR-10), Longformer concatenation takes 19ms on average to finish processing one image, while Linformer concatenation takes 7.50ms on a single RTX 2080ti GPU. This is because the sliding window mechanism in attention score calculation only calculates a few diagonals of a large matrix in matrix multiplication \citep{beltagy2020longformer}. This operation is not a traditional BLAS or Tensor operation, thus not optimized with CUDA kernels running on GPUs.

\begin{table}[t]
  \centering
  \begin{tabular}{lll}
    \toprule
    Attention Type     & Top1 Accuracy ($\%$)     & Top5 Accuracy ($\%$) \\
    \midrule
    Longformer Concatenation & 92.52  & 99.68     \\
    Linformer Concatenation  & 91.12 & 99.64      \\
    Longformer Replacement   & 89.86       & 99.55  \\
    Linformer Replacement     & 89.8       & 99.47  \\
    \bottomrule
  \end{tabular}
  \caption{We compare the performance of Concatenation or Replacement convolution blocks with Efficient Transformer Block. Using Longformer results in the best performance in both Concatenation and Replacement scenarios.}
\end{table}

\subsubsection{Number of Efficient Transformer Blocks}
Concatenating more attention layers produces more complex models and information. Therefore, we also test how augmenting more convolution layers with an Efficient Transformer Block affects the model's performance. Due to limited time and computation resources, we only test augmenting one or two layers of convolution blocks with Longformer Blocks. The results are shown in Table 3.

We find that adding more attention layers improves model performances. Moreover, we believe that if trained with higher resolution image datasets, the improvement will be more significant because the Efficient Transformer Block at deeper layers will have larger feature map input from convolution outputs.

\begin{table}[t]
  \centering
  \begin{tabular}{lll}
    \toprule
    Number of Concatenation     & Top1 Accuracy ($\%$)     & Top5 Accuracy ($\%$) \\
    \midrule
    1 & 91.14  & 99.58     \\
    2  & 92.52 & 99.68      \\
    \bottomrule
  \end{tabular}
  \caption{We compared augmenting effects with more Longformer Blocks. We find that concatenating more Longformer Blocks results in better model performance.}
\end{table}

\subsubsection{Downsampling Methods}
In ResNet architecture, a downsample process is done by either 2x2 max pooling or a convolution of stride two every couple of convolution layers \citep{he2016deep}. Because our augmented Efficient Transformer Block takes input from previous convolution layer output and concatenates it with downsampled convolution output of the current layer, Efficient Transformer Blocks also need to perform downsampling. There are two ways to achieve this:
\begin{itemize} 
    \item [1.] Using a stride two convolution layer before Efficient Transformer Block similar to ResNet and use 1x1 image patches for Efficient Transformer Blocks that do not change the input dimensions.
    \item [2.] Using 2x2 image patches for Efficient Transformer Blocks to reduce the input dimension by 2x.
\end{itemize}

We test the model performance of both approaches and show the results in Table 4. We find that using 2x2 image patches directly produces better performance. We think this is because using image patches larger than 1x1 pools more information into self-attention mechanism in Efficient Transformer Blocks

\begin{table}[t]
  \centering
  \begin{tabular}{lll}
    \toprule
    Downsampling Methods     & Top1 Accuracy ($\%$)     & Top5 Accuracy ($\%$) \\
    \midrule
    2x2 Image Patch & 90.90  & 99.56    \\
    Stride 2 Convolution  & 90.56 & 99.53      \\
    \bottomrule
  \end{tabular}
  \caption{We test the model performance using a 2x2 image patch for Efficient Transformer Blocks or a stride two convolution filter to match the downsampling procedure in ResNet. Using a 2x2 image patch produces the best model performance.}
\end{table}

\subsection{Training Complexity}
\subsubsection{Convergence}
We are interested in how difficult it is to train different approaches compared to the vanilla ResNet18. We plot the top1 accuracy of different models of every epoch until it converges in Figure 3. The Concatenation approach, in general, takes more epochs to converge than ResNet. This is expected as we added more parameters to learn the model.

The Replacement approach in our tests converges faster than vanilla Resnet. Theoretically, Transformer blocks with a self-attention mechanism should be more complex than convolution filters, and we expected the Replacement approach to take more epochs to converge. We think our test results are due to using CIFAR-10 where the image's resolution is 32x32. After a few downsampling processes, the dimension of input used in Transformer blocks is very small, making the model complexity similar to vanilla ResNet in the Replacement approach.

\begin{figure}[hbt!]
  \centering
  \includegraphics[scale=0.4]{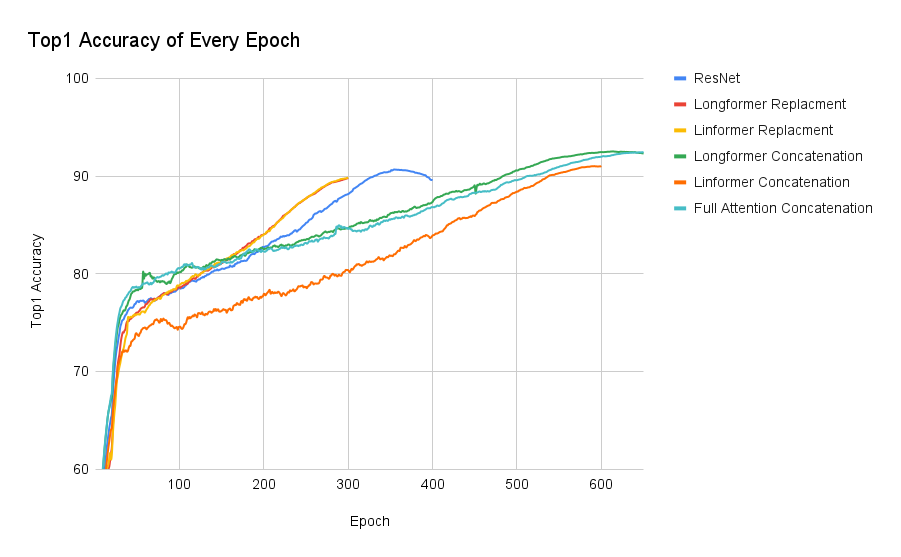}
  \caption{Top1 Accuracy of every Epoch until convergence. Longformer Replacement and Linformer Replacement Model converge faster than ResNet18. Longformer, Linformer, and Full Attention Concatenation converge at a similar number of epochs, with Linformer Concatenation being slightly faster. However, Longformer takes much more wall-clock time due to slow processing time.}
\end{figure}

\subsubsection{Peak Memory Usage}
The peak memory usage of our Efficient Transformer Augmentation model should scale linearly or close to linearly with input size compared to the quadratic scaling of augmenting normal Transformer blocks. We test this by processing images of different sizes (from 32x32 to 512x512 pixels with a batch size of 4) and measuring the peak memory usage in MBs. The results are shown in Figure 4. We find that augmenting with Linformer and Longformer scales much better than a normal transformer block, albeit not being strictly linear in practice. We have insufficient GPU memory (11GB VRAM on our testing GPU) to process 512x512 images. Our results show that Efficient Transformer Blocks can process high-resolution images or large feature maps that AA-Net with normal Transformers cannot work with.

\begin{figure}[hbt!]
  \centering
  \includegraphics[scale=0.4]{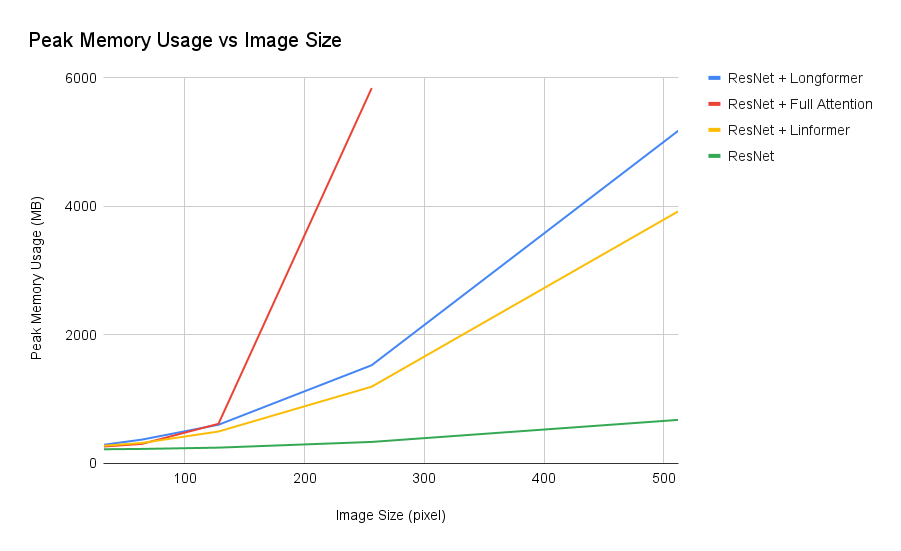}
  \caption{Peak Memory Usage of vanilla ResNet18 and ResNet18 adding different types of Transformer blocks. Longformer and Linformer scale better than normal Transformer blocks with larger input sizes.}
\end{figure}

\section{Limitations and Future Work}
Our model adapts from the work done on Multi-scale Vision Transformers \citep{zhang2021multi}; however, we did not fully adopt the position embedding interpolation logic. Our model at the current development stage disables the flexibility of the convolution network to fine-tune or infer different resolution images because the learned position embedding in Transformer blocks may not be meaningful for different image resolutions without interpolation. Future work might want to address this by adding interpolation methods for position embedding. 

Augmenting ResNet18 with Longformer suffers from significantly longer processing time due to the sliding window mechanism which results in not optimized matrix operations. In Multi-scale Vision Transformers, custom CUDA kernels are implemented for better optimization \citep{zhang2021multi}. Further work might investigate the effects of adopting these CUDA kernels.

Due to time and computational resource constraints, we only test model performance on CIFAR-10. Future work may also want to investigate the performance of our model on datasets with higher image resolution (e.g., ImageNet datasets). We think our model will have more performance gain on high-resolution image datasets because we will have more image patches and position embeddings in Efficient Transformer blocks, which should give the model more information.

\section{Conclusion}
In this work, we propose a novel approach to hybrid convolution-attention neural networks by adding Efficient Transformer blocks to convolution networks to improve the performance and alleviate high memory usage with high-resolution image input. We investigate two approaches to augmenting Transformer blocks to ResNet: Replacement or Concatenation. Our Longformer Concatenation model noticeably improves the performance of ResNet on CIFAR-10 while having better scalability with input size compared to the usage of traditional Transformer blocks with full attention compared to AA-Net.

% Implementation
\bibliography{ref}

\begin{thebibliography}{23}
\providecommand{\natexlab}[1]{#1}
\providecommand{\url}[1]{\texttt{#1}}
\expandafter\ifx\csname urlstyle\endcsname\relax
  \providecommand{\doi}[1]{doi: #1}\else
  \providecommand{\doi}{doi: \begingroup \urlstyle{rm}\Url}\fi

\bibitem[Bahdanau et~al.(2014)Bahdanau, Cho, and Bengio]{bahdanau2014neural}
D.~Bahdanau, K.~Cho, and Y.~Bengio.
\newblock Neural machine translation by jointly learning to align and
  translate.
\newblock \emph{arXiv preprint arXiv:1409.0473}, 2014.

\bibitem[Bello et~al.(2019)Bello, Zoph, Vaswani, Shlens, and
  Le]{bello2019attention}
I.~Bello, B.~Zoph, A.~Vaswani, J.~Shlens, and Q.~V. Le.
\newblock Attention augmented convolutional networks.
\newblock In \emph{Proceedings of the IEEE/CVF international conference on
  computer vision}, pages 3286--3295, 2019.

\bibitem[Beltagy et~al.(2020)Beltagy, Peters, and Cohan]{beltagy2020longformer}
I.~Beltagy, M.~E. Peters, and A.~Cohan.
\newblock Longformer: The long-document transformer.
\newblock \emph{arXiv preprint arXiv:2004.05150}, 2020.

\bibitem[Choromanski et~al.(2020)Choromanski, Likhosherstov, Dohan, Song, Gane,
  Sarlos, Hawkins, Davis, Mohiuddin, Kaiser, et~al.]{choromanski2020rethinking}
K.~Choromanski, V.~Likhosherstov, D.~Dohan, X.~Song, A.~Gane, T.~Sarlos,
  P.~Hawkins, J.~Davis, A.~Mohiuddin, L.~Kaiser, et~al.
\newblock Rethinking attention with performers.
\newblock \emph{arXiv preprint arXiv:2009.14794}, 2020.

\bibitem[Deng et~al.(2009)Deng, Dong, Socher, Li, Li, and
  Fei-Fei]{deng2009imagenet}
J.~Deng, W.~Dong, R.~Socher, L.-J. Li, K.~Li, and L.~Fei-Fei.
\newblock Imagenet: A large-scale hierarchical image database.
\newblock In \emph{2009 IEEE conference on computer vision and pattern
  recognition}, pages 248--255. Ieee, 2009.

\bibitem[He et~al.(2016)He, Zhang, Ren, and Sun]{he2016deep}
K.~He, X.~Zhang, S.~Ren, and J.~Sun.
\newblock Deep residual learning for image recognition.
\newblock In \emph{Proceedings of the IEEE conference on computer vision and
  pattern recognition}, pages 770--778, 2016.

\bibitem[Ho et~al.(2019)Ho, Kalchbrenner, Weissenborn, and
  Salimans]{ho2019axial}
J.~Ho, N.~Kalchbrenner, D.~Weissenborn, and T.~Salimans.
\newblock Axial attention in multidimensional transformers.
\newblock \emph{arXiv preprint arXiv:1912.12180}, 2019.

\bibitem[Hochreiter and Schmidhuber(1997)]{hochreiter1997long}
S.~Hochreiter and J.~Schmidhuber.
\newblock Long short-term memory.
\newblock \emph{Neural computation}, 9\penalty0 (8):\penalty0 1735--1780, 1997.

\bibitem[Hu et~al.(2018{\natexlab{a}})Hu, Shen, Albanie, Sun, and
  Vedaldi]{hu2018gather}
J.~Hu, L.~Shen, S.~Albanie, G.~Sun, and A.~Vedaldi.
\newblock Gather-excite: Exploiting feature context in convolutional neural
  networks.
\newblock \emph{Advances in neural information processing systems}, 31,
  2018{\natexlab{a}}.

\bibitem[Hu et~al.(2018{\natexlab{b}})Hu, Shen, and Sun]{hu2018squeeze}
J.~Hu, L.~Shen, and G.~Sun.
\newblock Squeeze-and-excitation networks.
\newblock In \emph{Proceedings of the IEEE conference on computer vision and
  pattern recognition}, pages 7132--7141, 2018{\natexlab{b}}.

\bibitem[Krizhevsky et~al.({\natexlab{a}})Krizhevsky, Nair, and
  Hinton]{cifar10}
A.~Krizhevsky, V.~Nair, and G.~Hinton.
\newblock Cifar-10 (canadian institute for advanced research).
\newblock {\natexlab{a}}.
\newblock URL \url{http://www.cs.toronto.edu/~kriz/cifar.html}.

\bibitem[Krizhevsky et~al.({\natexlab{b}})Krizhevsky, Nair, and
  Hinton]{cifar100}
A.~Krizhevsky, V.~Nair, and G.~Hinton.
\newblock Cifar-100 (canadian institute for advanced research).
\newblock {\natexlab{b}}.
\newblock URL \url{http://www.cs.toronto.edu/~kriz/cifar.html}.

\bibitem[Lee et~al.(2019)Lee, Lee, Kim, Kosiorek, Choi, and Teh]{lee2019set}
J.~Lee, Y.~Lee, J.~Kim, A.~Kosiorek, S.~Choi, and Y.~W. Teh.
\newblock Set transformer: A framework for attention-based
  permutation-invariant neural networks.
\newblock In \emph{International Conference on Machine Learning}, pages
  3744--3753. PMLR, 2019.

\bibitem[Park et~al.(2018)Park, Woo, Lee, and Kweon]{park2018bam}
J.~Park, S.~Woo, J.-Y. Lee, and I.~S. Kweon.
\newblock Bam: Bottleneck attention module.
\newblock \emph{arXiv preprint arXiv:1807.06514}, 2018.

\bibitem[Parmar et~al.(2018)Parmar, Vaswani, Uszkoreit, Kaiser, Shazeer, Ku,
  and Tran]{parmar2018image}
N.~Parmar, A.~Vaswani, J.~Uszkoreit, L.~Kaiser, N.~Shazeer, A.~Ku, and D.~Tran.
\newblock Image transformer.
\newblock In \emph{International Conference on Machine Learning}, pages
  4055--4064. PMLR, 2018.

\bibitem[Rabinovich et~al.(2007)Rabinovich, Vedaldi, Galleguillos, Wiewiora,
  and Belongie]{rabinovich2007objects}
A.~Rabinovich, A.~Vedaldi, C.~Galleguillos, E.~Wiewiora, and S.~Belongie.
\newblock Objects in context.
\newblock In \emph{2007 IEEE 11th International Conference on Computer Vision},
  pages 1--8. IEEE, 2007.

\bibitem[Rae et~al.(2019)Rae, Potapenko, Jayakumar, and
  Lillicrap]{rae2019compressive}
J.~W. Rae, A.~Potapenko, S.~M. Jayakumar, and T.~P. Lillicrap.
\newblock Compressive transformers for long-range sequence modelling.
\newblock \emph{arXiv preprint arXiv:1911.05507}, 2019.

\bibitem[Shaw et~al.(2018)Shaw, Uszkoreit, and Vaswani]{shaw2018self}
P.~Shaw, J.~Uszkoreit, and A.~Vaswani.
\newblock Self-attention with relative position representations.
\newblock \emph{arXiv preprint arXiv:1803.02155}, 2018.

\bibitem[Vaswani et~al.(2017)Vaswani, Shazeer, Parmar, Uszkoreit, Jones, Gomez,
  Kaiser, and Polosukhin]{vaswani2017attention}
A.~Vaswani, N.~Shazeer, N.~Parmar, J.~Uszkoreit, L.~Jones, A.~N. Gomez,
  {\L}.~Kaiser, and I.~Polosukhin.
\newblock Attention is all you need.
\newblock \emph{Advances in neural information processing systems}, 30, 2017.

\bibitem[Wang et~al.(2020)Wang, Li, Khabsa, Fang, and Ma]{wang2020linformer}
S.~Wang, B.~Z. Li, M.~Khabsa, H.~Fang, and H.~Ma.
\newblock Linformer: Self-attention with linear complexity.
\newblock \emph{arXiv preprint arXiv:2006.04768}, 2020.

\bibitem[Wang et~al.(2021)Wang, Xie, Li, Fan, Song, Liang, Lu, Luo, and
  Shao]{wang2021pyramid}
W.~Wang, E.~Xie, X.~Li, D.-P. Fan, K.~Song, D.~Liang, T.~Lu, P.~Luo, and
  L.~Shao.
\newblock Pyramid vision transformer: A versatile backbone for dense prediction
  without convolutions.
\newblock In \emph{Proceedings of the IEEE/CVF International Conference on
  Computer Vision}, pages 568--578, 2021.

\bibitem[Woo et~al.(2018)Woo, Park, Lee, and Kweon]{woo2018cbam}
S.~Woo, J.~Park, J.-Y. Lee, and I.~S. Kweon.
\newblock Cbam: Convolutional block attention module.
\newblock In \emph{Proceedings of the European conference on computer vision
  (ECCV)}, pages 3--19, 2018.

\bibitem[Zhang et~al.(2021)Zhang, Dai, Yang, Xiao, Yuan, Zhang, and
  Gao]{zhang2021multi}
P.~Zhang, X.~Dai, J.~Yang, B.~Xiao, L.~Yuan, L.~Zhang, and J.~Gao.
\newblock Multi-scale vision longformer: A new vision transformer for
  high-resolution image encoding.
\newblock In \emph{Proceedings of the IEEE/CVF International Conference on
  Computer Vision}, pages 2998--3008, 2021.

\end{thebibliography}
\bibliographystyle{abbrvnat}

\appendix
\end{document}